\pdfoutput=1

\documentclass[11pt]{article}

\usepackage[]{ACL2023}

\usepackage{times}
\usepackage{latexsym}
\usepackage{cleveref}

\usepackage[T1]{fontenc}

\usepackage[utf8]{inputenc}

\usepackage{microtype}

\usepackage{adjustbox}
\usepackage{graphicx}
\usepackage{booktabs}
\usepackage{multirow}
\usepackage{float}

\RequirePackage[inline,shortlabels]{enumitem}
\setlist{nosep}

\RequirePackage[textsize=footnotesize, color=yellow!20]{todonotes}

\usepackage{inconsolata}

\newcommand{\dataset}{mOKB6}
\newcommand{\mono}{{\scshape Mono}}
\newcommand{\union}{{\scshape Union}}
\newcommand{\trans}{{\scshape Trans}}
\newcommand{\monotrans}{{\scshape Mono+Trans}}
\newcommand{\uniontrans}{{\scshape Union+Trans}}
\newcommand{\moiemodel}{{\scshape Gen2OIE}}
\newcommand{\unionnoen}{{\scshape Union} w/o En}
\newcommand{\memorize}{{\scshape Memorize}}

\newcommand\blfootnote[1]{%
  \begingroup
  \renewcommand\thefootnote{}\footnote{#1}%
  \addtocounter{footnote}{-1}%
  \endgroup
}

\def\ztitle{\dataset{}: A Multilingual Open Knowledge Base Completion Benchmark}
\title{\ztitle}

\author{
  Shubham Mittal\textsuperscript{$\alpha\dagger$} \hspace{1em}
  Keshav Kolluru\textsuperscript{$\beta\dagger$} \hspace{1em}
  Soumen Chakrabarti\textsuperscript{$\gamma$} \hspace{1em}
  Mausam\textsuperscript{$\alpha$} \\
  \textsuperscript{$\alpha$} Indian Institute of Technology Delhi \\ 
  \textsuperscript{$\beta$} KnowDis AI, New Delhi \\    
  \textsuperscript{$\gamma$} Indian Institute of Technology Bombay \\
  \texttt{shubhamiitd18@gmail.com, keshav.kolluru@gmail.com} \\
  \texttt{soumen@cse.iitb.ac.in, mausam@cse.iitd.ac.in}
}

\begin{document}
\maketitle
\blfootnote{$\dagger$ Major part of work done as students at IIT Delhi.}

\begin{abstract}
Automated completion of open knowledge bases (Open KBs), which are constructed from triples of the form (\textit{subject phrase}, \textit{relation phrase}, \textit{object phrase}), obtained via open information extraction (Open IE) system, are useful for discovering novel facts that may not be directly present in the text.
However, research in Open KB completion (Open KBC) has so far been limited to resource-rich languages like English.
Using the latest advances in multilingual Open IE, we construct the first multilingual Open KBC dataset, called \dataset{}, containing facts from Wikipedia in six languages (including English).
Improving the previous Open KB construction pipeline by doing multilingual coreference resolution and keeping only entity-linked triples, we create a \textit{dense} Open KB.
We experiment with several models for the task and observe a consistent benefit of combining languages with the help of shared embedding space as well as translations of facts.
We also observe that current multilingual models struggle to remember facts seen in languages of different scripts.\footnote{Dataset and code released at \href{https://github.com/dair-iitd/mokb6.git}{github.com:dair-iitd/mokb6}}
\end{abstract}

\section{Introduction}
\label{sec:intro}
Open information extraction (Open IE) systems \cite{mausam16} such as ReVerb \cite{etzioni11} and OpenIE6 \cite{kolluru20openie6} can extract triples, or \textit{facts}, of the form (\textit{subject phrase}, \textit{relation phrase}, \textit{object phrase}), which can be denoted as $(s,r,o)$, from text (e.g., Wikipedia articles) without using any pre-defined ontology. 
Open knowledge base (Open KB) is constructed using these Open IE triples where the subject phrases and object phrases are nodes and relation phrases are labels on edges connecting the nodes in the graph.
Open knowledge base completion (Open KBC) is the task of discovering new links between nodes using the graph structure of the Open KB.
Knowledge graph embedding (KGE) models are typically used for the Open KBC task, where they are asked to answer questions of the form $(s,r,?)$ and $(?,r,o)$.

Research in Open KBC has been restricted to English \cite{cesi} due to lack of Open KBs in other languages. 
We aim to study multilingual Open KBC, with the motivation that the information available in high resource languages like English may help when inferring links in Open KBs that use low resource languages like Telugu.  
Moreover, intuitively, if all the information in different languages can be pooled together, then it may help the model learn better, and allow information flow across Open KBs in different languages.

We design the first multilingual Open KB construction pipeline (shown in \Cref{fig:pipeline}) using a multilingual Open IE system, \moiemodel{} \cite{gen2oie}.
We find that coreference resolution is missing in existing Open KB construction \cite{opiec} but is important for increasing the coverage of facts (as described in \Cref{fig:corefImp}).
We re-train a recent coref model \cite{wlcoref} using XLM-R \cite{xlmr} as the underlying multilingual encoder and add it to our pipeline.
For constructing a high quality test set, we use 988 manually verified facts in English. 
For extending to other languages, we automatically translate English facts. 
The dataset thus constructed, called \dataset{}, contains 42K facts in six languages: English, Hindi, Telugu, Spanish, Portuguese, and Chinese.


We report the first baselines for multilingual Open KBC task. 
We find that they are able to benefit from information in multiple languages when compared to using facts from a single language. 
Translations of Open KB facts also help the models.
However, we notice that although the multilingual KGE models learn facts in a particular language, they struggle to remember the same fact, when queried in another language with different script.


\section{Related Work}
\label{sec:related}
Multilingual Open KBC datasets are absent in literature to the best of our knowledge, although multiple English Open KBC datasets are available. 
OLPBench \cite{olpbench}, derived from OPIEC \cite{opiec}, is a large-scale Open KBC dataset that contains 30M triples and is constructed from English Wikipedia using MinIE system \cite{minie}.
The evaluation data contains 10K triples randomly sampled from 1.25M \textit{linked} triples.
ReVerb45K \cite{cesi} and ReVerb20K \cite{reverb20K} are smaller Open KBC datasets constructed from Clueweb09 corpus\footnote{\url{http://www.lemurproject.org/clueweb09.php/}} using ReVerb Open IE system \cite{reverbIE}.
Both the datasets keep only those tuples in which both the \emph{subject phrase} and \emph{object phrase} link to a finite set of Freebase entities. 

Multilingual Open IE (mOpenIE) systems like \moiemodel{} \cite{gen2oie} and Multi$^2$OIE \cite{multi2oie} enable extracting facts from multiple languages.
We use the \moiemodel{} model for constructing \dataset{} dataset as it is trained with language-specific facts transferred from English, while Multi$^2$OIE relies on zero-shot transfer for languages other than English.

\paragraph{Knowledge Graph Embedding (KGE) Models:} 
Conventional KGE models like TransE \cite{transe}, ComplEx \cite{complex}, ConvE \cite{conve}, and TuckER \cite{tucker} have been used for Open KBC task \cite{care, olpbench, okgit, kbcMeetsTransferLearning}.
Given a triple $(s,r,o)$, these models encode the \textit{subject phrase}, \textit{relation phrase}, and \textit{object phrase} from free text, and pass the encodings to a triple-scoring function, which is optimized using binary cross entropy loss. ComplEx has also been used for multilingual closed KBC task \cite{chakrabarti-etal-2022-joint}. 



Pretrained language models like BERT \cite{bert} have been used in KGE models for the KBC task \cite{framework-plm-kge, plm-kge-eval, okgit, kgbert}. 
SimKGC \cite{simkgc} is the state of the art KGE model on closed KBC task. 
It computes the score of a triple $(s,r,o)$ as the cosine similarity of the embeddings of $(s; r)$ and $(o)$, computed using two separate pretrained BERT models without any weight sharing.

\section{Dataset Curation} 
\label{sec:data_curation}
\begin{figure*}[htb!]
\centering
\includegraphics[width=\textwidth]{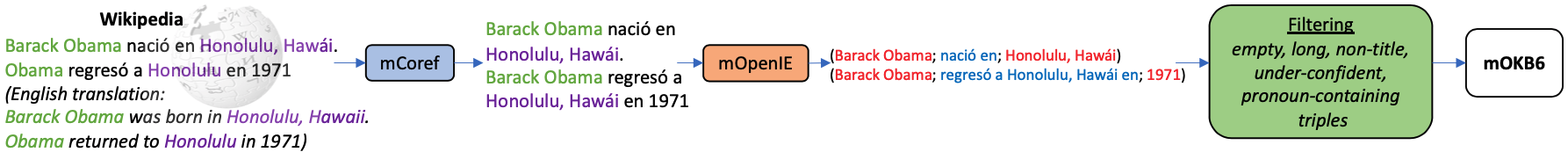} 
\caption{Our three-staged multilingual Open KB construction pipeline for \dataset{}. 
mCoref is multilingual coreference resolution system, having XLM-R \cite{xlmr} encoder based wl-coref \cite{wlcoref}, and mOpenIE is multilingual open information extraction system, consisting of \moiemodel{} \cite{gen2oie}.}
\label{fig:pipeline}
\vspace{-1ex}
\end{figure*}

We aim to construct a \textit{dense} multilingual Open KB that maximizes the information about a given real-world entity, which may be represented as multiple nodes across languages.
Therefore, we consider those Wikipedia articles\footnote{Wikidump of April 02, 2022} that are available in six languages: English, Hindi, Telugu, Spanish, Portuguese, and Chinese\footnote{languages are chosen to match availability of Gen2OIE}.
This will also help the model learn from facts in high resource language like English and answer queries in low resource language like Telugu.
We work with 300 titles randomly sampled from the ones common among all six languages (found using MediaWiki-Langlinks \cite{mediawiki}).
Thus, we extract facts from 6${\times}$300 Wikipedia articles.
We discuss the three stages of our pipeline below.

\paragraph{Stage 1} We first process each Wikipedia article through a coreference resolution system.
Although language-specific end-to-end neural coref models have been developed \cite{crac22-multilingual-crf, movingontonotes}, multilingual models that work on all our languages of interest are absent in the literature.
Therefore, we retrain wl-coref \cite{wlcoref} with XLM-R \cite{xlmr} on the English training data (available in OntoNotes \cite{ontonotes}) that can work zero-shot for other languages.

Coref models detect and cluster mentions, but do not identify a canonical cluster name, which is needed for standardizing all the mentions in the cluster.
To find cluster names, entity linking systems such as mGENRE \cite{mgenre} or Wikipedia hyperlinks can be used. 
However, we found that they result in low recall, particularly for low resource languages. 
Thus, we employ a heuristic to find the cluster name and replace each of the coreferent mentions with it. 
The score for each mention is represented by a tuple, computed as: Score(mention phrase) = (\#proper nouns, \#nouns, \#numerals, \# adjectives, \#pronouns, \#verbs).
The tuple is ordered according to the importance of each field (POS tags) for the cluster name, which is determined empirically.
Two tuples are compared index-wise with higher priority given to lower indices to determine the best scoring mention that is chosen as the canonical name (\Cref{tab:mentionscorer}).

\begin{table}[h]
\small
\centering
\begin{tabular}{@{}llll@{}}
\toprule
Mentions & Scores & Cluster Name \\ \midrule
Barack Obama & \textbf{(2,0,0,0,0,0)} &  \\
Obama & (1,0,0,0,0,0) & \multirow{1}{*}{Barack Obama} \\
He & (0,0,0,0,1,0) & \\ \bottomrule
\end{tabular}
\caption{Parts of speech tags are used to find the canoniccal name of the coreferent cluster of entity mentions.} 
\label{tab:mentionscorer}
\vspace{-3ex}
\end{table}

\paragraph{Stage 2} We use \moiemodel{} to extract Open IE triples from the coreference resolved sentences.

\paragraph{Stage 3} Similar to \citet{opiec}, we apply various filters to remove \textit{noisy} triples that have empty or very long arguments, or have less confidence than 0.3 (as assigned by \moiemodel{}). 
We further only keep triples that have the article's title as either the \textit{subject phrase} or \textit{object phrase}, to avoid generic or specific triples, valid only in the particular context.
Examples of \textit{contextual} triples \cite{choi-etal-2021-decontextualization} are discussed in \Cref{app:contextual}.
See \Cref{app:pipeline_stats} for further data curation details.

\begin{figure}[h!]
\centering 
\includegraphics[scale=0.3]{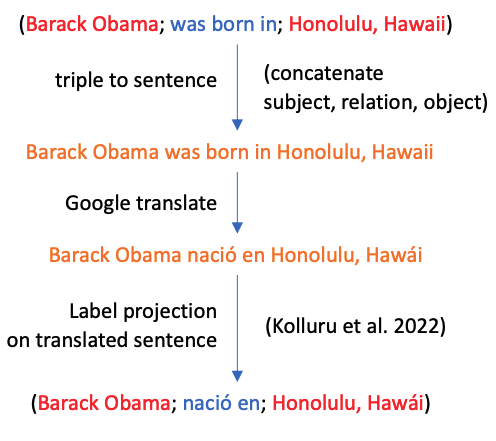} 
\caption{Method to translate Open IE triple using Google translate, and followed by label projection using word alignments \cite{gen2oie}.}
\label{fig:transproject}
\end{figure}

These automatically extracted triples form the train set of \dataset{}.
To form a high quality test set in six languages with limited access to experts in all languages, the test set is created in a semi-automatic way.
We sample 1600 English triples from the train set (which are subsequently filtered) and manually remove noisy triples. 
We use inter-annotation agreement between two annotators to check if they both agree that the given triple is noisy or clean.
With an agreement of 91\%, we retain 988 English triples, which we automatically translate to the other five languages.
As illustrated in \Cref{fig:transproject}, to translate a triple, we convert it to a sentence after removing tags and use Google translate\footnote{\url{https://translate.google.co.in/}} for translating the triple-converted sentence to the remaining five languages.
We observed high quality of translated triples, with 88\% satisfactory translations as determined by native-speakers of three languages on a set of 75 translated triples.
To get the Open IE \textit{subject phrase}, \textit{relation phrase} and \textit{object phrase} tags, we project the labels from the original English triple to the translated sentence using word alignments \cite{gen2oie}.
Finally, we are left with 550 triples in each language after removing examples where some labels could not be aligned. 
We use these 6${\times}$550 triples as the test sets.
The train and dev sets are created from the remaining triples in each language such that the dev set has 500 randomly sampled triples (\Cref{tab:datastats}). 

We analyse the entity overlap across languages and find that on an average, a test entity (which is present in either the \textit{subject phrase} or \textit{object phrase} of a test tuple) is present 17.73 times in English, 0.94 times in Hindi, 0.47 times in Telugu, 2.33 times in Spanish, 1.69 times in Portuguese, and 1.45 times in Chinese train set.

Our construction pipeline improves over OPIEC in three ways:
\begin{enumerate*}[(1)]
\item we use a multilingual Open IE system, instead of an English-specific Open IE system like in OPIEC, enabling us to curate Open KBs in many languages,
\item we add a multilingual coreference resolution system in our pipeline, and
\item the English test triples are manually verified.
\end{enumerate*}
Further, we manually evaluate and review the noise at each step of data curation in \Cref{sec:evalnoise}.

\begin{table}[h!]
\small
\centering
\begin{tabular}{@{}llccccc@{}}
\toprule
                 & En    & Hi   & Te   & Es   & Pt   & Zh   \\ \midrule
\#entity         & 20637 & 4625 & 3972 & 5651 & 5304 & 5037 \\
\#relation       & 7870  & 2177 & 1907 & 2823 & 2644 & 2325 \\
\#train          & 20195 & 2786 & 1992 & 3966 & 3528 & 3420 \\ \bottomrule
\end{tabular}
\caption{Statistics of individual Open KBs in \dataset{} in English (En), Hindi (Hi), Telugu (Te), Spanish (Es), Portuguese (Pt), and Chinese (Zh). The dev and test set for each Open KB contain 500 and 550 triples each.}
\label{tab:datastats}
\end{table}

\section{Noise Evaluation}
\label{sec:evalnoise}
Curating an Open KB involves various stages and each stage induces its noise in the construction pipeline \cite{opiec}. 
We manually evaluate the noise induced at each stage of our pipeline (\Cref{fig:pipeline}) and discuss the same in this section.
We ask native speakers of four (out of six) languages - English, Hindi, Telugu, and Chinese to assess the output quality, or precision, of each stage as discussed below.

In the first stage, we assess the performance of the coreference resolution system over Wikipedia articles.
We find a high precision of 95.5\% in coref's mention clustering and 89.82\% accuracy in finding canonical cluster name (using the heuristic illustrated in \Cref{tab:mentionscorer}), computed over 40 randomly sampled coref clusters (10 in each language). 

For evaluating the Open IE system, \moiemodel{}, in the second stage, we mark an extraction of a sentence as correct if it has syntactically correct arguments and it is coherent with the sentence. 
We get an average precision of 63.4\% on 80 extractions (20 in each language).

We evaluate the triples, or Open KB facts, at the last stage after passing through various noise-removing filters. 
Note that these triples also form the train set (and dev set) in \dataset{} dataset.
We mark triples as correct when they contain real-world entities, and also, factual information about them.
If the triple is very generic or contextual (see \Cref{app:contextual}), it is marked as incorrect.
We find the train (and dev) set quality to be 69.3\%, averaged over 80 triples in four languages.


\section{Experiments}
\label{sec:exps}

\begin{table*}[htp!]
\adjustbox{max size=\hsize}{ \tabcolsep 3pt
\begin{tabular}{l|rrr|rrr|rrr|rrr|rrr|rrr} \hline
 & \multicolumn{3}{c|}{English (En)} & \multicolumn{3}{c|}{Hindi (Hi)} & \multicolumn{3}{c|}{Telugu (Te)} & \multicolumn{3}{c|}{Spanish (Es)} & \multicolumn{3}{c|}{Portuguese (Pt)} & \multicolumn{3}{c}{Chinese (Zh)} \\
& {\tiny H@1}  & {\tiny H@10}   & {\tiny MRR}   & {\tiny H@1}   & {\tiny H@10}   & {\tiny MRR}   & {\tiny H@1}   & {\tiny H@10}   & {\tiny MRR}   & {\tiny H@1}   & {\tiny H@10}   & {\tiny MRR}   & {\tiny H@1}   & {\tiny H@10}   & {\tiny MRR}   & {\tiny H@1}   & {\tiny H@10}   & {\tiny MRR}   \\ \hline 
\mono{}                 & 14.8   & 38.7   & 22.8  & 3.0    & 14.8   & 7.2   & 1.5    & 8.1    & 3.9   & 6.4    & 23.7   & 12.3  & 6.3    & 21.7   & 11.4  & 2.4    & 13.1  & 6.2     \\
\unionnoen{}    & 5.7   & 21.5   & 10.9  & 2.9    & 15.4   & 7.4   & 1.8    & 10.2    & 4.9   & 8.1    & 27.8   & 14.5  & 6.7    & 26.1   & 12.9  & 3.2    & 15.5  & 7.5    \\
\union{}    & \textbf{16.7}     & \textbf{40.8}   & \textbf{24.8}  & 3.6    & 16.6   & 8.1   & 1.5    & 9.3   & 4.5   & 10.6    & 32.2   & 17.6  & 9.7   & 29.3   & 16.6  & 4.0    & 18.8  & 8.9   \\
\trans{}       & -   & -     & -  & 20.5   & 47.6   & 29.7  & 8.7   & 28.7   & 15.5  & 23.2   & 50.6   & 32.4    & 20.5  & 50.7   & 30.5  & 14.0   & 39.4  & 22.5  \\
\monotrans{}       & -   & -     & -  & 20.2   & 45.4   & 28.4  & 14.3  & \textbf{38.5}   & 22.2  & 23.5    & 51.5   & 32.9    & 21.4   & 48.9   & 30.7  & \textbf{17.9}   & 43.2  & \textbf{26.6}  \\ 
\uniontrans{}   & -   & -     & -  & \textbf{23.3}   & \textbf{49.7}   & \textbf{32.3}  & \textbf{15.1}   & \textbf{38.5}   & \textbf{23.1}  & \textbf{23.9}   & \textbf{52.4}   & \textbf{33.4}    &  \textbf{23.5}   & \textbf{52.1}   & \textbf{33.1}  & 16.9   & \textbf{43.6}  & 26.0  \\ \hline
\end{tabular}}
\caption{Performance (\%) of SimKGC model on \dataset{} dataset, comprising of Open KBs in six languages. \mono{}, \trans{}, and \monotrans{} are monolingual models trained only on facts of one language whereas \union{}, \unionnoen{}, and \uniontrans{} are multilingual models trained with facts from multiple languages. All reported numbers are an average of three runs using different seeds. Best scores are highlighted in bold.}
\label{tab:finalresults}
\end{table*}

Our experimental study on multilingual open KBC task investigates the following research questions:
\begin{enumerate}
    \item Does the KGE model benefit from facts in different languages? (\Cref{subsec:q1})
    \item Can translation help transfer among languages? (\Cref{subsec:q3})
    \item Does the KGE model remember facts seen across different languages? (\Cref{subsec:q2})
\end{enumerate}

We use SimKGC model \citep{simkgc} with pretrained mBERT initialization to run our experiments, after comparing with recent KGE models (\Cref{app:comparekgemodels}). 
For evaluation, we use three metrics --- hits at rank~1 (H@1), hits at rank~10 (H@10), and mean reciprocal rank (MRR).
The formal definitions of them are provided in \Cref{app:metrics}.
We discuss further model training details in \Cref{app:trainingdetails}.

\subsection{Training on Multilingual Facts}
\label{subsec:q1}
We train and compare monolingual model, called \mono{}, with multilingual models, \union{} and \unionnoen{}.
In \mono{}, we train one model for each language using its respective Open KB, whereas in \union{}, a single model is trained on six languages' Open KBs together.
\union{} outperforms \mono{} in all languages by an average of 4.6\% H@10 and 2.8\% MRR (see \Cref{tab:finalresults}), which provides evidence of information flow across languages and the model benefits from it. 

To check the extent of flow from (high-resource) English to the other languages, we also train on the five languages except English, which we call \unionnoen{}.
We find \unionnoen{} also outperforms \mono{} by 2.7\% H@10 and 1.2\% MRR over the five languages, hinting that interlingual transfer is more general and pervasive.

\subsection{Open KB Facts Translation}
\label{subsec:q3}

Apart from relying only on multilingual transfer in the embedding space, we analyse the effect of using translated triples in the training of the KGE model.
We translate the English training triples\footnote{English source achieved the best translation quality.} to the other five languages (\Cref{sec:data_curation})
and train monolingual models using only the translated triples (\trans{}).
To leverage facts present in each language's Open KB, we make \monotrans{}, where we add language-specific \mono{} data to the translated triples.
\Cref{tab:finalresults} shows that \monotrans{} is better than \mono{} by a large margin of 15.5\% H@1, 29.2\% H@10, and 20.0\% MRR, averaged over five languages.
Also, \monotrans{} improves over \trans{} by 2.1\% H@10 and 2.0\% MRR, showcasing the importance of facts in each language's Open KBs.

To effectively gain from transfer in both the embedding space as well as translation, we introduce \uniontrans{}. 
We train one model for each language, on the combination of \union{} triples and the translated train triples from English Open KB to that language.
\uniontrans{} is better than \union{} by 25.9\% H@10 and 18.4\% MRR.
This suggests that the model is able to benefit from English facts when they are translated to the query language, unlike in \union{} where the English facts are present only in English.

\subsection{Cross-lingual Memorization}
\label{subsec:q2}
\begin{figure}[h!]
\centering 
\includegraphics[scale=0.45]{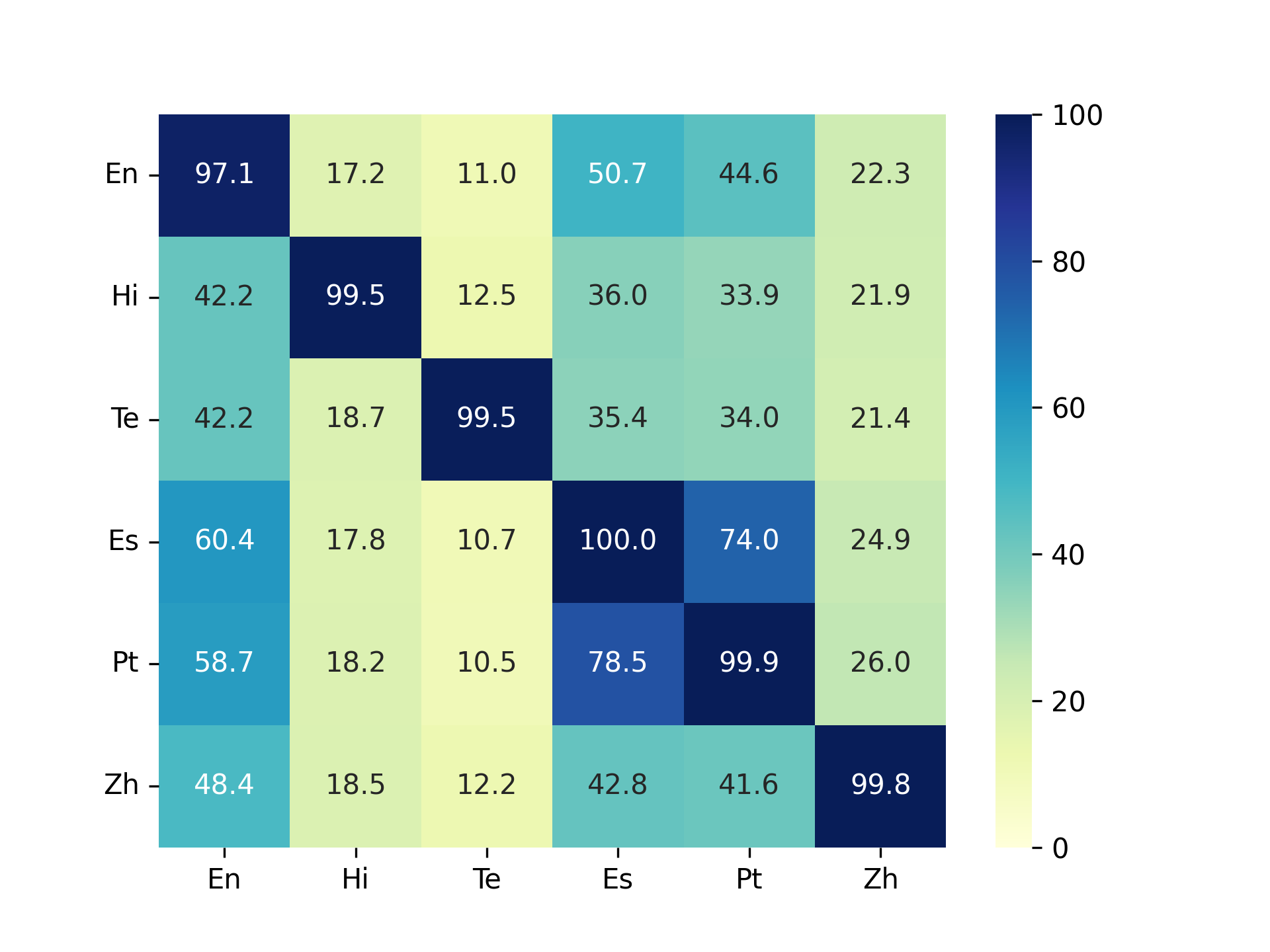} 
\caption{Performance (H@10) of \memorize{} models. Row $L$ shows the performance of \memorize{}$_L$ model across the test sets of all languages (columns). For example, the performance of \memorize{}$_{En}$ when tested on English (En) is 97.1\% H@10, and \memorize{}$_{En}$ when tested on Spanish (Es) gives 50.7\% H@10. We find relatively good cross-lingual transfer among languages that use same script (Latin in English, Spanish and Portuguese) compared to those using different scripts (English, Hindi, Telugu and Chinese).}
\vspace{-2ex}
\label{fig:memorize}
\end{figure}

Pretrained multilingual language models such as mBERT have demonstrated strong cross-lingual transfer capabilities \cite{suprisingCrosslingual}.
We investigate cross-lingual memorization of the KGE model by showing facts in one language and querying the same facts in other five languages. 
For each language, $L$, we take the \union{} model and train it further on the test set of that language's Open KB, which we call \memorize{}$_L$ model.
Then, we test each \memorize{}$_L$ model on the six test sets.
Since the test sets (in \dataset{} dataset) of the different languages contain the same facts, this experiment allows us to investigate cross-lingual memorization.
We provide the H@10 scores of \memorize{} models in \Cref{fig:memorize} and the performance on other metrics (H@1 and MRR) is reported in \Cref{tab:memorize}.

The model achieves at least 97\% H@10 when tested on the language used for training (diagonal). 
We observe that there is relatively good cross-lingual memorization among languages that share the same script (Latin in English, Spanish, and Portuguese), but the model struggles to remember facts when seen in languages of different scripts. 
Many entities look similar in shared scripts, possibly leading to better information transfer. 
For example, the \memorize{}$_{En}$ achieves H@10 of 50.7\% in Spanish (Es) compared to 22.3\% in Chinese (Zh) and 11\% in Telugu (Te).

\section{Conclusion and Future Work}
\label{sec:conclusion}
We create and release the \dataset{} dataset, the first multilingual Open Knowledge Base Completion dataset with 42K facts in six languages: English, Hindi, Telugu, Spanish, Portuguese, and Chinese.
Its construction uses multilingual coreference resolution, entity-mention cluster naming, multilingual open information extraction and various filtering steps to improve the quality of the extracted facts.
We also report the first baselines on the task using the existing state of the art KGE models trained with facts from different languages using various augmentation strategies.

Our work opens many important research questions: (1) Can we develop better strategies to combine facts in different languages? (2) Can we build models that achieve strong information transfer across unrelated languages with same or different scripts? (3) Can we train the neural model to ignore contextual triples (\Cref{app:contextual}), thus improving overall performance? and (4) Can tying the same entities across various languages help the model generalize better?
We leave these questions to be addressed in future work.

\section{Acknowledgements}
\label{sec:acknowledgements}
Keshav was supported by TCS Research Fellowship during his PhD. 
Mausam is supported by grants from Huawei, Google, Verisk and IBM, and a Jai Gupta Chair Fellowship. He also acknowledges Google and Yardi School of AI travel grants. 
Soumen is partly supported by a Jagadish Bose Fellowship and a grant from Cisco. 
We thank IIT Delhi HPC facility for compute resources. 

\section{Limitations}
Although multilingual, the constructed open KB is limited to the sampling of the chosen six languages.  We do not know how well the system will generalize to various language families that have not been considered here.
Further, even among the languages considered, the performance of even the best-performing systems, as measured through H@1 is still in the low 20's.
Therefore the models are not yet ready to be deployed for real-world applications.

\bibliography{anthology,custom}

\begin{thebibliography}{39}
\expandafter\ifx\csname natexlab\endcsname\relax\def\natexlab#1{#1}\fi

\bibitem[{Balazevic et~al.(2019)Balazevic, Allen, and Hospedales}]{tucker}
Ivana Balazevic, Carl Allen, and Timothy Hospedales. 2019.
\newblock \href {https://doi.org/10.18653/v1/D19-1522} {{T}uck{ER}: Tensor
  factorization for knowledge graph completion}.
\newblock In \emph{Proceedings of the 2019 Conference on Empirical Methods in
  Natural Language Processing and the 9th International Joint Conference on
  Natural Language Processing (EMNLP-IJCNLP)}, pages 5185--5194, Hong Kong,
  China. Association for Computational Linguistics.

\bibitem[{Bordes et~al.(2013)Bordes, Usunier, Garcia-Duran, Weston, and
  Yakhnenko}]{transe}
Antoine Bordes, Nicolas Usunier, Alberto Garcia-Duran, Jason Weston, and Oksana
  Yakhnenko. 2013.
\newblock \href
  {https://proceedings.neurips.cc/paper/2013/file/1cecc7a77928ca8133fa24680a88d2f9-Paper.pdf}
  {Translating embeddings for modeling multi-relational data}.
\newblock In \emph{Advances in Neural Information Processing Systems},
  volume~26. Curran Associates, Inc.

\bibitem[{Broscheit et~al.(2020)Broscheit, Gashteovski, Wang, and
  Gemulla}]{olpbench}
Samuel Broscheit, Kiril Gashteovski, Yanjie Wang, and Rainer Gemulla. 2020.
\newblock \href {https://doi.org/10.18653/v1/2020.acl-main.209} {Can we predict
  new facts with open knowledge graph embeddings? a benchmark for open link
  prediction}.
\newblock In \emph{Proceedings of the 58th Annual Meeting of the Association
  for Computational Linguistics}, pages 2296--2308, Online. Association for
  Computational Linguistics.

\bibitem[{Chakrabarti et~al.(2022)Chakrabarti, Singh, Lohiya, Jain, and
  ~}]{chakrabarti-etal-2022-joint}
Soumen Chakrabarti, Harkanwar Singh, Shubham Lohiya, Prachi Jain, and Mausam ~.
  2022.
\newblock \href {https://aclanthology.org/2022.emnlp-main.817} {Joint
  completion and alignment of multilingual knowledge graphs}.
\newblock In \emph{Proceedings of the 2022 Conference on Empirical Methods in
  Natural Language Processing}, pages 11922--11938, Abu Dhabi, United Arab
  Emirates. Association for Computational Linguistics.

\bibitem[{Chandrahas and Talukdar(2021)}]{okgit}
.~Chandrahas and Partha Talukdar. 2021.
\newblock \href {https://doi.org/10.18653/v1/2021.findings-acl.225} {{OKGIT}:
  {O}pen knowledge graph link prediction with implicit types}.
\newblock In \emph{Findings of the Association for Computational Linguistics:
  ACL-IJCNLP 2021}, pages 2546--2559, Online. Association for Computational
  Linguistics.

\bibitem[{Cho et~al.(2014)Cho, van Merri{\"e}nboer, Gulcehre, Bahdanau,
  Bougares, Schwenk, and Bengio}]{gru}
Kyunghyun Cho, Bart van Merri{\"e}nboer, Caglar Gulcehre, Dzmitry Bahdanau,
  Fethi Bougares, Holger Schwenk, and Yoshua Bengio. 2014.
\newblock \href {https://doi.org/10.3115/v1/D14-1179} {Learning phrase
  representations using {RNN} encoder{--}decoder for statistical machine
  translation}.
\newblock In \emph{Proceedings of the 2014 Conference on Empirical Methods in
  Natural Language Processing ({EMNLP})}, pages 1724--1734, Doha, Qatar.
  Association for Computational Linguistics.

\bibitem[{Choi et~al.(2021)Choi, Palomaki, Lamm, Kwiatkowski, Das, and
  Collins}]{choi-etal-2021-decontextualization}
Eunsol Choi, Jennimaria Palomaki, Matthew Lamm, Tom Kwiatkowski, Dipanjan Das,
  and Michael Collins. 2021.
\newblock \href {https://doi.org/10.1162/tacl_a_00377} {Decontextualization:
  Making sentences stand-alone}.
\newblock \emph{Transactions of the Association for Computational Linguistics},
  9:447--461.

\bibitem[{Conneau et~al.(2020)Conneau, Khandelwal, Goyal, Chaudhary, Wenzek,
  Guzman, Grave, Ott, Zettlemoyer, and Stoyanov}]{xlmr}
Alexis Conneau, Kartikay Khandelwal, Naman Goyal, Vishrav Chaudhary, Guillaume
  Wenzek, Francisco Guzman, Edouard Grave, Myle Ott, Luke Zettlemoyer, and
  Veselin Stoyanov. 2020.
\newblock \href {https://doi.org/10.18653/v1/2020.acl-main.747} {Unsupervised
  cross-lingual representation learning at scale}.
\newblock In \emph{{ACL} Conference}, pages 8440--8451.

\bibitem[{De~Cao et~al.(2022)De~Cao, Wu, Popat, Artetxe, Goyal, Plekhanov,
  Zettlemoyer, Cancedda, Riedel, and Petroni}]{mgenre}
Nicola De~Cao, Ledell Wu, Kashyap Popat, Mikel Artetxe, Naman Goyal, Mikhail
  Plekhanov, Luke Zettlemoyer, Nicola Cancedda, Sebastian Riedel, and Fabio
  Petroni. 2022.
\newblock \href {https://doi.org/10.1162/tacl_a_00460} {Multilingual
  autoregressive entity linking}.
\newblock \emph{Transactions of the Association for Computational Linguistics},
  10:274--290.

\bibitem[{Dettmers et~al.(2018)Dettmers, Minervini, Stenetorp, and
  Riedel}]{conve}
Tim Dettmers, Pasquale Minervini, Pontus Stenetorp, and Sebastian Riedel. 2018.
\newblock \href {https://ojs.aaai.org/index.php/AAAI/article/view/11573/11432}
  {Convolutional 2d knowledge graph embeddings}.
\newblock In \emph{Proceedings of the Thirty-Second AAAI Conference on
  Artificial Intelligence and Thirtieth Innovative Applications of Artificial
  Intelligence Conference and Eighth AAAI Symposium on Educational Advances in
  Artificial Intelligence}, AAAI'18/IAAI'18/EAAI'18. AAAI Press.

\bibitem[{Devlin et~al.(2019)Devlin, Chang, Lee, and Toutanova}]{bert}
Jacob Devlin, Ming-Wei Chang, Kenton Lee, and Kristina Toutanova. 2019.
\newblock \href {https://doi.org/10.18653/v1/N19-1423} {{BERT}: Pre-training of
  deep bidirectional transformers for language understanding}.
\newblock In \emph{Proceedings of the 2019 Conference of the North {A}merican
  Chapter of the Association for Computational Linguistics: Human Language
  Technologies, Volume 1 (Long and Short Papers)}, pages 4171--4186,
  Minneapolis, Minnesota. Association for Computational Linguistics.

\bibitem[{Dobrovolskii(2021)}]{wlcoref}
Vladimir Dobrovolskii. 2021.
\newblock \href {https://aclanthology.org/2021.emnlp-main.605} {Word-level
  coreference resolution}.
\newblock In \emph{Proceedings of the 2021 Conference on Empirical Methods in
  Natural Language Processing}, pages 7670--7675, Online and Punta Cana,
  Dominican Republic. Association for Computational Linguistics.

\bibitem[{Etzioni et~al.(2011)Etzioni, Fader, Christensen, Soderland, and
  Mausam}]{etzioni11}
Oren Etzioni, Anthony Fader, Janara Christensen, Stephen Soderland, and Mausam.
  2011.
\newblock \href {https://www.ijcai.org/Proceedings/11/Papers/012.pdf} {Open
  information extraction: The second generation}.
\newblock In \emph{{IJCAI} 2011, Proceedings of the 22nd International Joint
  Conference on Artificial Intelligence, Barcelona, Catalonia, Spain, July
  16-22, 2011}, pages 3--10. {IJCAI/AAAI}.

\bibitem[{Fader et~al.(2011)Fader, Soderland, and Etzioni}]{reverbIE}
Anthony Fader, Stephen Soderland, and Oren Etzioni. 2011.
\newblock \href {https://aclanthology.org/D11-1142} {Identifying relations for
  open information extraction}.
\newblock In \emph{Proceedings of the 2011 Conference on Empirical Methods in
  Natural Language Processing}, pages 1535--1545, Edinburgh, Scotland, UK.
  Association for Computational Linguistics.

\bibitem[{Gal\'{a}rraga et~al.(2014)Gal\'{a}rraga, Heitz, Murphy, and
  Suchanek}]{reverb20K}
Luis Gal\'{a}rraga, Geremy Heitz, Kevin Murphy, and Fabian~M. Suchanek. 2014.
\newblock \href {https://doi.org/10.1145/2661829.2662073} {Canonicalizing open
  knowledge bases}.
\newblock New York, NY, USA. Association for Computing Machinery.

\bibitem[{Gashteovski et~al.(2017)Gashteovski, Gemulla, and del Corro}]{minie}
Kiril Gashteovski, Rainer Gemulla, and Luciano del Corro. 2017.
\newblock \href {https://doi.org/10.18653/v1/D17-1278} {{M}in{IE}: Minimizing
  facts in open information extraction}.
\newblock In \emph{Proceedings of the 2017 Conference on Empirical Methods in
  Natural Language Processing}, pages 2630--2640, Copenhagen, Denmark.
  Association for Computational Linguistics.

\bibitem[{Gashteovski et~al.(2019)Gashteovski, Wanner, Hertling, Broscheit, and
  Gemulla}]{opiec}
Kiril Gashteovski, Sebastian Wanner, Sven Hertling, Samuel Broscheit, and
  Rainer Gemulla. 2019.
\newblock \href {https://openreview.net/pdf?id=HJxeGb5pTm} {Opiec: An open
  information extraction corpus}.
\newblock In \emph{Proceedings of the Conference on Automatic Knowledge Base
  Construction (AKBC)}.

\bibitem[{Gupta et~al.(2019)Gupta, Kenkre, and Talukdar}]{care}
Swapnil Gupta, Sreyash Kenkre, and Partha Talukdar. 2019.
\newblock \href {https://doi.org/10.18653/v1/D19-1036} {{C}a{R}e: Open
  knowledge graph embeddings}.
\newblock In \emph{Proceedings of the 2019 Conference on Empirical Methods in
  Natural Language Processing and the 9th International Joint Conference on
  Natural Language Processing (EMNLP-IJCNLP)}, pages 378--388, Hong Kong,
  China. Association for Computational Linguistics.

\bibitem[{Kim et~al.(2020)Kim, Hong, Ko, and Seo}]{kgbert}
Bosung Kim, Taesuk Hong, Youngjoong Ko, and Jungyun Seo. 2020.
\newblock \href {https://doi.org/10.18653/v1/2020.coling-main.153} {Multi-task
  learning for knowledge graph completion with pre-trained language models}.
\newblock In \emph{Proceedings of the 28th International Conference on
  Computational Linguistics}, pages 1737--1743, Barcelona, Spain (Online).
  International Committee on Computational Linguistics.

\bibitem[{Kingma and Ba(2015)}]{adam}
Diederik~P. Kingma and Jimmy Ba. 2015.
\newblock \href {http://arxiv.org/abs/1412.6980} {Adam: {A} method for
  stochastic optimization}.
\newblock In \emph{3rd International Conference on Learning Representations,
  {ICLR} 2015, San Diego, CA, USA, May 7-9, 2015, Conference Track
  Proceedings}.

\bibitem[{Kocijan and Lukasiewicz(2021)}]{kbcMeetsTransferLearning}
Vid Kocijan and Thomas Lukasiewicz. 2021.
\newblock \href {https://aclanthology.org/2021.emnlp-main.524.pdf} {Knowledge
  base completion meets transfer learning}.
\newblock In \emph{Proceedings of the 2021 Conference on Empirical Methods in
  Natural Language Processing (EMNLP)}, Punta Cana, Dominican Republic.
  Association for Computational Linguistics.

\bibitem[{Kolluru et~al.(2020)Kolluru, Adlakha, Aggarwal, {Mausam}, and
  Chakrabarti}]{kolluru20openie6}
Keshav Kolluru, Vaibhav Adlakha, Samarth Aggarwal, {Mausam}, and Soumen
  Chakrabarti. 2020.
\newblock \href {https://doi.org/10.18653/v1/2020.emnlp-main.306} {{O}pen{IE}6:
  {I}terative {G}rid {L}abeling and {C}oordination {A}nalysis for {O}pen
  {I}nformation {E}xtraction}.
\newblock In \emph{Proceedings of the 2020 Conference on Empirical Methods in
  Natural Language Processing (EMNLP)}, pages 3748--3761, Online. Association
  for Computational Linguistics.

\bibitem[{Kolluru et~al.(2022)Kolluru, Mohammed, Mittal, Chakrabarti, and
  .}]{gen2oie}
Keshav Kolluru, Muqeeth Mohammed, Shubham Mittal, Soumen Chakrabarti, and
  Mausam . 2022.
\newblock \href {https://doi.org/10.18653/v1/2022.acl-long.179}
  {Alignment-augmented consistent translation for multilingual open information
  extraction}.
\newblock In \emph{Proceedings of the 60th Annual Meeting of the Association
  for Computational Linguistics (Volume 1: Long Papers)}, pages 2502--2517,
  Dublin, Ireland. Association for Computational Linguistics.

\bibitem[{Lovelace and Ros{\'e}(2022)}]{framework-plm-kge}
Justin Lovelace and Carolyn Ros{\'e}. 2022.
\newblock \href {https://aclanthology.org/2022.emnlp-main.398} {A framework for
  adapting pre-trained language models to knowledge graph completion}.
\newblock In \emph{Proceedings of the 2022 Conference on Empirical Methods in
  Natural Language Processing}, pages 5937--5955, Abu Dhabi, United Arab
  Emirates. Association for Computational Linguistics.

\bibitem[{Lv et~al.(2022)Lv, Lin, Cao, Hou, Li, Liu, Li, and
  Zhou}]{plm-kge-eval}
Xin Lv, Yankai Lin, Yixin Cao, Lei Hou, Juanzi Li, Zhiyuan Liu, Peng Li, and
  Jie Zhou. 2022.
\newblock \href {https://doi.org/10.18653/v1/2022.findings-acl.282} {Do
  pre-trained models benefit knowledge graph completion? a reliable evaluation
  and a reasonable approach}.
\newblock In \emph{Findings of the Association for Computational Linguistics:
  ACL 2022}, pages 3570--3581, Dublin, Ireland. Association for Computational
  Linguistics.

\bibitem[{Mausam(2016)}]{mausam16}
Mausam. 2016.
\newblock \href {https://www.ijcai.org/Proceedings/16/Papers/604.pdf} {Open
  information extraction systems and downstream applications}.
\newblock In \emph{International Joint Conference on Artificial Intelligence}.

\bibitem[{MediaWiki(2021)}]{mediawiki}
MediaWiki. 2021.
\newblock \href
  {https://www.mediawiki.org/w/index.php?title=API:Langlinks&oldid=4936309}
  {Api:langlinks --- mediawiki{,}}.
\newblock [Online; accessed 02-April-2022].

\bibitem[{Pennington et~al.(2014)Pennington, Socher, and Manning}]{glove}
Jeffrey Pennington, Richard Socher, and Christopher Manning. 2014.
\newblock \href {https://doi.org/10.3115/v1/D14-1162} {{G}lo{V}e: Global
  vectors for word representation}.
\newblock In \emph{Proceedings of the 2014 Conference on Empirical Methods in
  Natural Language Processing ({EMNLP})}, pages 1532--1543, Doha, Qatar.
  Association for Computational Linguistics.

\bibitem[{Qi et~al.(2020)Qi, Zhang, Zhang, Bolton, and Manning}]{stanza}
Peng Qi, Yuhao Zhang, Yuhui Zhang, Jason Bolton, and Christopher~D. Manning.
  2020.
\newblock \href {https://nlp.stanford.edu/pubs/qi2020stanza.pdf} {Stanza: A
  {Python} natural language processing toolkit for many human languages}.
\newblock In \emph{Proceedings of the 58th Annual Meeting of the Association
  for Computational Linguistics: System Demonstrations}.

\bibitem[{Ro et~al.(2020)Ro, Lee, and Kang}]{multi2oie}
Youngbin Ro, Yukyung Lee, and Pilsung Kang. 2020.
\newblock \href {https://doi.org/10.18653/v1/2020.findings-emnlp.99}
  {{M}ulti{\^{}}2{OIE}: Multilingual open information extraction based on
  multi-head attention with {BERT}}.
\newblock In \emph{Findings of the Association for Computational Linguistics:
  EMNLP 2020}, pages 1107--1117, Online. Association for Computational
  Linguistics.

\bibitem[{Trouillon et~al.(2016)Trouillon, Welbl, Riedel, Gaussier, and
  Bouchard}]{complex}
Théo Trouillon, Johannes Welbl, Sebastian Riedel, Eric Gaussier, and Guillaume
  Bouchard. 2016.
\newblock \href {https://proceedings.mlr.press/v48/trouillon16.html} {Complex
  embeddings for simple link prediction}.
\newblock In \emph{Proceedings of The 33rd International Conference on Machine
  Learning}, volume~48 of \emph{Proceedings of Machine Learning Research},
  pages 2071--2080, New York, New York, USA. PMLR.

\bibitem[{Vashishth et~al.(2018)Vashishth, Jain, and Talukdar}]{cesi}
Shikhar Vashishth, Prince Jain, and Partha Talukdar. 2018.
\newblock \href {https://doi.org/10.1145/3178876.3186030} {{CESI}:
  Canonicalizing open knowledge bases using embeddings and side information}.
\newblock In \emph{Proceedings of the 2018 World Wide Web Conference}, WWW '18,
  pages 1317--1327, Republic and Canton of Geneva, Switzerland. International
  World Wide Web Conferences Steering Committee.

\bibitem[{Wang et~al.(2022)Wang, Zhao, Wei, and Liu}]{simkgc}
Liang Wang, Wei Zhao, Zhuoyu Wei, and Jingming Liu. 2022.
\newblock \href {https://doi.org/10.18653/v1/2022.acl-long.295} {{S}im{KGC}:
  Simple contrastive knowledge graph completion with pre-trained language
  models}.
\newblock In \emph{Proceedings of the 60th Annual Meeting of the Association
  for Computational Linguistics (Volume 1: Long Papers)}, pages 4281--4294,
  Dublin, Ireland. Association for Computational Linguistics.

\bibitem[{Weischedel et~al.(2013)Weischedel, Palmer, Marcus, Hovy, Pradhan,
  Ramshaw, Xue, Taylor, Kaufman, and Franchini}]{ontonotes}
Ralph Weischedel, Martha Palmer, Mitchell Marcus, Eduard Hovy, Sameer Pradhan,
  Lance Ramshaw, Nianwen Xue, Ann Taylor, Jeff Kaufman, and Michelle Franchini.
  2013.
\newblock \href {https://doi.org/10.35111/xmhb-2b84} {Ontonotes release 5.0}.
\newblock In \emph{Linguistic Data Consortium, Philadelphia, PA}.

\bibitem[{Wenzek et~al.(2020)Wenzek, Lachaux, Conneau, Chaudhary, Guzm{\'a}n,
  Joulin, and Grave}]{commoncrawl}
Guillaume Wenzek, Marie-Anne Lachaux, Alexis Conneau, Vishrav Chaudhary,
  Francisco Guzm{\'a}n, Armand Joulin, and Edouard Grave. 2020.
\newblock \href {https://aclanthology.org/2020.lrec-1.494} {{CCN}et: Extracting
  high quality monolingual datasets from web crawl data}.
\newblock In \emph{Proceedings of the Twelfth Language Resources and Evaluation
  Conference}, pages 4003--4012, Marseille, France. European Language Resources
  Association.

\bibitem[{Wu and Dredze(2019)}]{suprisingCrosslingual}
Shijie Wu and Mark Dredze. 2019.
\newblock \href {https://doi.org/10.18653/v1/D19-1077} {Beto, bentz, becas: The
  surprising cross-lingual effectiveness of {BERT}}.
\newblock In \emph{Proceedings of the 2019 Conference on Empirical Methods in
  Natural Language Processing and the 9th International Joint Conference on
  Natural Language Processing (EMNLP-IJCNLP)}, pages 833--844, Hong Kong,
  China. Association for Computational Linguistics.

\bibitem[{Xia and Van~Durme(2021)}]{movingontonotes}
Patrick Xia and Benjamin Van~Durme. 2021.
\newblock \href {https://doi.org/10.18653/v1/2021.emnlp-main.425} {Moving on
  from {O}nto{N}otes: Coreference resolution model transfer}.
\newblock In \emph{Proceedings of the 2021 Conference on Empirical Methods in
  Natural Language Processing}, pages 5241--5256, Online and Punta Cana,
  Dominican Republic. Association for Computational Linguistics.

\bibitem[{Xue et~al.(2021)Xue, Constant, Roberts, Kale, Al-Rfou, Siddhant,
  Barua, and Raffel}]{mt5}
Linting Xue, Noah Constant, Adam Roberts, Mihir Kale, Rami Al-Rfou, Aditya
  Siddhant, Aditya Barua, and Colin Raffel. 2021.
\newblock \href {https://doi.org/10.18653/v1/2021.naacl-main.41} {m{T}5: A
  massively multilingual pre-trained text-to-text transformer}.
\newblock In \emph{Proceedings of the 2021 Conference of the North American
  Chapter of the Association for Computational Linguistics: Human Language
  Technologies}, pages 483--498, Online. Association for Computational
  Linguistics.

\bibitem[{{\v{Z}}abokrtsk{\'y} et~al.(2022){\v{Z}}abokrtsk{\'y}, Konop{\'\i}k,
  Nedoluzhko, Nov{\'a}k, Ogrodniczuk, Popel, Pra{\v{z}}{\'a}k, Sido, Zeman, and
  Zhu}]{crac22-multilingual-crf}
Zden{\v{e}}k {\v{Z}}abokrtsk{\'y}, Miloslav Konop{\'\i}k, Anna Nedoluzhko,
  Michal Nov{\'a}k, Maciej Ogrodniczuk, Martin Popel, Ond{\v{r}}ej
  Pra{\v{z}}{\'a}k, Jakub Sido, Daniel Zeman, and Yilun Zhu. 2022.
\newblock \href {https://aclanthology.org/2022.crac-mcr.1} {Findings of the
  shared task on multilingual coreference resolution}.
\newblock In \emph{Proceedings of the CRAC 2022 Shared Task on Multilingual
  Coreference Resolution}, pages 1--17, Gyeongju, Republic of Korea.
  Association for Computational Linguistics.

\end{thebibliography}
\bibliographystyle{acl_natbib}

\appendix

\clearpage

\twocolumn[\centering \Large\bfseries \ztitle\\ \large (Appendix) \\ \vspace{2ex} ]

\section{Dataset Curation}
\label{app:pipeline_stats}

As discussed in \Cref{sec:data_curation}, we construct \dataset{} dataset in three stages after extracting the Wikipedia articles (using WikiExtractor\footnote{\url{https://github.com/samuelbroscheit/wikiextractor-wikimentions}}) from the Wikidump of April 02, 2022.
We run our construction pipeline (as shown in \Cref{fig:pipeline}) for all six languages on a single V100 (32 GB) GPU, which required 14 hours of computation to create \dataset{} dataset.

In the first stage, we keep the sentences containing at least 6 and at most 50 tokens since we find that most of the short sentences are headings or sub-headings present in Wikipedia articles, and very long sentences can't be input to \moiemodel{} (in second stage) due to maximum sequence length constraint of 1024 in mT5 \cite{mt5} based \moiemodel{}.
This filtering step discards 18.9\% of sentences on an average in all six languages.
We use Stanza \cite{stanza} to perform sentence- and word-segmentation on Wikipedia articles in all six languages.
After filtering the sentences, the articles are processed for coreference resolution using XLM-R \cite{xlmr} encoder based wl-coref \cite{wlcoref}, followed by replacing the coreferent cluster mentions with their canonical cluster name using the heuristic discussed in \Cref{sec:data_curation}.

In the second stage, the coreference resolved articles are passed through \moiemodel{} to get the Open IE triples.
The confidence scores for these triples are computed using label rescoring, for which we refer the readers to \citet{gen2oie} for more details. 

Finally, in the last stage, we apply various filters, adapted from \citet{opiec}, to remove triples that are of no interest to Open KBC task, like the triples:
\begin{enumerate*}[(1)]
\item having any of its argument or relation empty,
\item containing more than 10 tokens in any of its arguments or relation,
\item having confidence score less than 0.3,
\item containing pronouns (found using Stanza) in its arguments,
\item having same subject and object (i.e. self loops), and
\item that are duplicates.
\end{enumerate*}
These filters keep 91.6\% of the triples obtained from stage 2 in all six languages. \\

Further in the last stage, in order to create a \textit{dense} Open KB containing minimum noise and maximum facts about the entities, we keep the triples having the Wikipedia article's title as either the \textit{subject phrase} or \textit{object phrase} and discard the rest.
We do this by finding all the coreference clusters (of entity mentions) that contain the titles, then get the entities, or cluster names, of those clusters using the heuristic discussed in \cref{sec:data_curation}, and keep those triples that contain these cluster names.
This filtering step retains 23.6\% of the triples.

\section{Metrics}
\label{app:metrics}
We follow the previous works \cite{simkgc} on the evaluation methodology of Open KBC task and apply it to the multilingual Open KBC task, containing facts in multiple languages.
Given an Open KB, containing a finite set of entities and open relations, the KGE model answers forward and backward queries of the form $(s,r,?)$ and $(?,r,o)$ respectively.
The model ranks all the entities based on their correctness with, say, $s$ and $r$ in the forward query.
Further, the evaluation is in \textit{filtered} setting, where the other known correct answers, apart from $o$, are removed from  rank list.

The commonly used evaluation metrics are hits at rank~N (H@N), where $N$ is a natural number, and mean reciprocal rank (MRR).
Suppose, the model ranks $o$ at $R$ among all entities.
Then, H@N measures how many times $R$ is less than or equal to $N$. 
MRR is the average of reciprocal ranks ($\frac{1}{R}$). 
Both, H@N and MRR, are computed as average over both forms of queries over the full test set.

\begin{figure*}
\centering
\includegraphics[width=\textwidth]{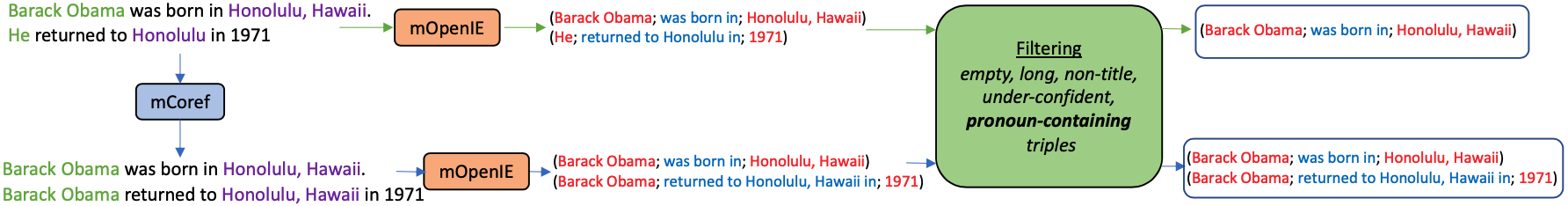} 
\caption{Previous Open KB construction pipelines like \citet{opiec} (shown by green arrows) lack coreference resolution system, which result in filtering \textit{important} facts like (\textit{Barack Obama}; \textit{returned to Honolulu, Hawaii in}; \textit{1971}). Our pipeline (shown by blue arrows) increases the \textit{coverage} of facts due to mCoref system.}
\label{fig:corefImp}
\end{figure*}

\section{Knowledge Graph Embedding Models}
\label{app:comparekgemodels}
SimKGC \cite{simkgc} is a text-based KGE model that uses two unshared pretrained BERT models \cite{bert} for encoding \textit{(subject phrase; relation phrase)} and \textit{object phrase} separately.
GRU-ConvE \cite{kbcMeetsTransferLearning} encodes both the \textit{relation phrase} and \textit{argument phrase} from their surface forms using two unshared GRU \cite{gru}.
CaRe \cite{care} learns separate embeddings for each \textit{argument phrase} and uses a bi-directional GRU to encode the \textit{relation phrase} from its surface form.
Both, GRU-ConvE and CaRe, are initialised with Glove embeddings \cite{glove}. 

To choose the best model for our experiments (\Cref{tab:finalresults}, \Cref{fig:memorize}), we train the recent knowledge graph embedding (KGE) models --- CaRe,, GRU-ConvE and SimKGC on the English Open KB in \dataset{}.
We report performance in \Cref{tab:threeKGEs} using the three metrics: hits at rank~1 (H@1), hits at~10 (H@10), and mean reciprocal rank (MRR).
We find that SimKGC with BERT encoder outperforms the other two models.

\begin{table}[h!]
\small
\centering
\begin{tabular}{@{}lccc@{}}
\toprule
          & H@1        & H@10       & MRR           \\ \midrule
CaRe     & 6.6           & 11.3           & 8.3           \\
GRU-ConvE & 12.4          & 27.8          & 17.8          \\
SimKGC (BERT) & \textbf{16.1} & \textbf{40.0} & \textbf{24.3} \\ \midrule
SimKGC (mBERT)  & \textbf{14.8} & \textbf{38.7} & \textbf{22.8} \\
SimKGC (XLM-R)  & 13.8 & 35.8 & 21.3 \\ \bottomrule
\end{tabular}
\caption{Performance (\%) of the KGE models on the English test set in \dataset{} dataset. The reported numbers are an average of three runs using different seeds.}
\label{tab:threeKGEs}
\end{table}

Since BERT supports only English language, we replace BERT in SimKGC with multilingual pre-trained language models like mBERT \cite{bert} or XLM-R \cite{xlmr}, to extend SimKGC model to other languages.
We find in \Cref{tab:threeKGEs} that SimKGC with mBERT is better than with XLM-R by 2.9\% H@10 and 1.5\% MRR, possibly because mBERT (and \dataset{}) uses Wikipedia while XLM-R uses CommonCrawl \cite{commoncrawl} during pre-training.
Thus, we use SimKGC with mBERT as the underlying encoder to run our experiments for all the languages.

\section{KGE Model Training Details}
\label{app:trainingdetails}
We use the code from official repositories of the KGE models --- SimKGC \cite{simkgc}, GRU-ConvE \cite{kbcMeetsTransferLearning}, and CaRe \cite{care} for our experiments.
The models are trained using Adam optimizer \cite{adam} on a single A100 (40 GB) GPU with three different random seeds and we report the average of three evaluation runs.

We do not perform hyperparameter search trials, except for batch size, and use the default hyperparameters from the respective codes of KGE models (see \Cref{tab:kge_hyperparams}).
We use early stopping to find the best model checkpoints based on HITS@1.
The dev set is different for each baseline: \mono{}, \trans{}, \monotrans{}, and \uniontrans{} use individual language's dev set, whereas \unionnoen{} and \union{} use the English dev set.
We report the performance of baseline models on the dev sets in \Cref{tab:dev_results_1} and \Cref{tab:dev_results_2}.

\begin{table}[h!]
\small
\centering
\begin{tabular}{@{}llll@{}}
\toprule
Hyperparameter    & SimKGC  & GRU-ConvE & CaRe \\ \midrule
\#epochs          & 100      & 500        & 500   \\
\#patience epochs & 10      & 10        & 10   \\
learning rate     & 3e-5    & 3e-4      & 1e-3 \\
dropout           & 0.1     & 0.3       & 0.5  \\
batch size        & 256 & 1024      & 128  \\
additive margin   & 0.02    & N/A       & N/A  \\ \bottomrule
\end{tabular}
\caption{Hyperparameters of the KGE models.}
\label{tab:kge_hyperparams}
\end{table}

We provide the number of trainable parameters of each KGE model in \Cref{tab:num_model_params_time}.
Based on the batch size and model size, different experiments consume different GPU hours.
To train on English Open KB (in \dataset{} dataset), CaRe and GRU-ConvE models took 2.5 hours and 0.5 hours, respectively, whereas SimKGC takes nearly 1 hour of GPU time.

\begin{table}[h!]
\small
\centering
\begin{tabular}{@{}ll@{}}
\toprule
KGE model       & \#trainable parameters  \\ \midrule
CaRe            & 12,971,423    \\
GRU-ConvE       & 12,085,523    \\
SimKGC (BERT)   & 216,620,545   \\
SimKGC (mBERT)  & 355,706,881   \\
SimKGC (XLM-R)  & 1,119,780,865 \\ \bottomrule
\end{tabular}
\caption{Number of trainable parameters in the KGE models.}
\label{tab:num_model_params_time}
\end{table}

\begin{table*}[htp!]
\adjustbox{max size=\hsize}{ \tabcolsep 3pt
\begin{tabular}{l|rrr|rrr|rrr|rrr|rrr|rrr} \hline
 & \multicolumn{3}{c|}{English} & \multicolumn{3}{c|}{Hindi} & \multicolumn{3}{c|}{Telugu} & \multicolumn{3}{c|}{Spanish} & \multicolumn{3}{c|}{Portuguese} & \multicolumn{3}{c}{Chinese} \\
& {\tiny H@1}  & {\tiny H@10}   & {\tiny MRR}   & {\tiny H@1}   & {\tiny H@10}   & {\tiny MRR}   & {\tiny H@1}   & {\tiny H@10}   & {\tiny MRR}   & {\tiny H@1}   & {\tiny H@10}   & {\tiny MRR}   & {\tiny H@1}   & {\tiny H@10}   & {\tiny MRR}   & {\tiny H@1}   & {\tiny H@10}   & {\tiny MRR}   \\ \hline 
English                 & 68.4   & 97.1     & 78.8  & 3.4    & 17.2     & 8.3     & 1.6    & 11      & 5     & 17.8   & 50.7     & 28.6  & 17     & 44.6   & 26    & 5.4    & 22.3  & 11.1  \\
Hindi                   & 19     & 42.2     & 26.7  & 80.6   & 99.5     & 88.3    & 2.4    & 12.5    & 5.9   & 12.3   & 36       & 19.9  & 12.3   & 33.9   & 19.7  & 5.3    & 21.9  & 10.8  \\ 
Telugu                  & 19.5   & 42.2     & 27.2  & 4.3    & 18.7     & 9.4     & 74.4   & 99.5    & 84.2  & 10.9   & 35.4     & 18.9  & 10.7   & 34   & 18.5  & 4.7   & 21.4    & 10.1  \\ 
Spanish                 & 27.9   & 60.4     & 38.8  & 4.1    & 17.8     & 8.9     & 1.8    & 10.7    & 5.1   & 84     & 100      & 90.3  & 37.6   & 74     & 50.1  & 6.5    & 24.9  & 12.8  \\ 
Portuguese              & 27.8   & 58.7     & 38.2  & 4.4    & 18.2     & 9.3     & 1.7    & 10.5    & 5.1   & 41.5   & 78.5     & 53.6  & 84.2   & 99.9   & 90.8  & 6.6    & 26    & 13.2  \\ 
Chinese                 & 22.1   & 48.4      & 30.6  & 3.5    & 18.5     & 8.8     & 1.8    & 12.2    & 5.4   & 14.8   & 42.8     & 24.2  & 15.7   & 41.6   & 24.1  & 81.6   & 99.8  & 89.2  \\ \hline
\end{tabular}}
\caption{Performance (\%) of the six \memorize{} models, which have been trained on each language's test set and tested on all the test sets in \dataset{} dataset.}
\label{tab:memorize}
\end{table*}

\newpage

\section{Contextual Triples}
\label{app:contextual}
Open IE triples are of various kinds and not all of them can be used for Open KBC task.
Various filtering steps are used to remove some of these in data curation (\Cref{sec:data_curation}).
We define \textit{contextual} triples as another kind of noisy triples, which are specific to, and are not interpretable out of, the context of text from which they are extracted.

\begin{table}[h!]
\small
\centering
\begin{tabular}{c}
\hline
(\textit{Max Born}; \textit{continued}; \textit{scientific work}) \\
(\textit{Robb Gravett}; \textit{won}; \textit{the championship}) \\
(\textit{George Herbert Walker Bush}; \textit{was}; \textit{out of touch}) \\
(\textit{Christianity}; \textit{is}; \textit{dominant}) \\ \hline
\end{tabular}
\caption{Examples of contextual triples.}
\label{tab:contextualtable}
\end{table}

From the first two triples in \Cref{tab:contextualtable}, it is unclear which scientific work \textit{Max Born} continued, or which championship \textit{Robb Gravett} has won.
The last two triples are too specific to the context and contain no factual information.

\begin{table*}[htp!]
\adjustbox{max size=\hsize}{ \tabcolsep 3pt
\begin{tabular}{l|rrr|rrr|rrr|rrr|rrr|rrr} \hline
 & \multicolumn{3}{c|}{English (En)} & \multicolumn{3}{c|}{Hindi (Hi)} & \multicolumn{3}{c|}{Telugu (Te)} & \multicolumn{3}{c|}{Spanish (Es)} & \multicolumn{3}{c|}{Portuguese (Pt)} & \multicolumn{3}{c}{Chinese (Zh)} \\
& {\tiny H@1}  & {\tiny H@10}   & {\tiny MRR}   & {\tiny H@1}   & {\tiny H@10}   & {\tiny MRR}   & {\tiny H@1}   & {\tiny H@10}   & {\tiny MRR}   & {\tiny H@1}   & {\tiny H@10}   & {\tiny MRR}   & {\tiny H@1}   & {\tiny H@10}   & {\tiny MRR}   & {\tiny H@1}   & {\tiny H@10}   & {\tiny MRR}   \\ \hline 
\mono{}                 & 16.2   & 38.7   & 23.9  & 18.2    & 39.4   & 25.9   & 8.5    & 20     & 12.5   & 17.3    & 36.6   & 23.7    & 17.6    & 39.6   & 25.3  & 10.8  & 31.9  & 17.8     \\
\trans{}                & -      & -      & -     & 8.1     & 23.7   & 13.5   & 3.3    & 15.4   & 7.5    & 12.9    & 33.6   & 20.3    & 12.6    & 37.2   & 20.6  & 5     & 20.8  & 10.3     \\
\monotrans{}            & -      & -      & -     & 20.8    & 43.2   & 28.6   & 7.8    & 24.8   & 13.4   & 20.2    & 46     & 28.8    & 21      & 45.9   & 29.2  & 10.6  & 30.1  & 16.7     \\
\union{}                & 19.9   & 39.6   & 26.4  & 14.5     & 38.2   & 22.4    & 5.9   & 20   & 10.6   & 19.8    & 43.2   & 27.9    & 19.7    & 43.8   & 28     & 11.2  & 33    & 18.8     \\
\unionnoen{}            & 5.8    & 19.5   & 10.6  & 15.4     & 39.3   & 23.3    & 6.3   & 20.5 & 11.1   & 19.4    & 41.6   & 26.4    & 16.9    & 42.9   & 25.9   & 11.3  & 33    & 18.4     \\
\uniontrans{}           & -      & -      & -     & 20.8    & 44.9   & 28.8   & 7.3    & 27.1   & 14    & 21.4   & 45.3    & 29.6     & 19.4    & 49.1    & 29.1 & 6.9   & 31    & 15.1     \\ \hline
\end{tabular}}
\caption{Performance (\%) of SimKGC on the dev sets (of \dataset{} dataset) in six languages.}
\label{tab:dev_results_1}
\end{table*}

\begin{table*}[h!]
\small
\centering
\begin{tabular}{@{}lccc@{}}
\toprule
                & H@1        & H@10       & MRR           \\ \midrule
CaRe            & 7.1        & 11.1       & 8.5           \\
GRU-ConvE       & 16.8       & 31.5       & 22.1          \\
SimKGC (BERT)   & 20.3       & 40.1       & 27.1          \\ \midrule
SimKGC (mBERT)  & 16.2       & 38.7       & 23.9          \\
SimKGC (XLM-R)  & 17         & 36.6       & 23.2          \\ \bottomrule
 
\end{tabular}
\caption{Performance (\%) of the KGE models on dev set of English Open KB in \dataset{} dataset.}
\label{tab:dev_results_2}
\end{table*}


\end{document}